\documentclass[11pt,a4paper]{article}
\usepackage[hyperref]{acl2018}
\usepackage{latexsym}

\usepackage{times}
\usepackage{xcolor}
\usepackage{natbib}
\usepackage{soul}
\usepackage[utf8]{inputenc}
\usepackage[small]{caption}

\usepackage{CJKutf8}
\usepackage{url}

\usepackage{amsmath,amsthm,amssymb,amsfonts,bbm,stmaryrd}
\usepackage{enumitem}

\usepackage{multirow} 
\usepackage{graphicx} 
\usepackage{booktabs} 
\usepackage{subcaption}
\usepackage[normalem]{ulem}
\usepackage{makecell}
\usepackage{courier}
\usepackage{bbm}
\usepackage{algorithm,algpseudocode}
\usepackage{adjustbox}
\usepackage{tabularx}
\usepackage{booktabs}
\usepackage{setspace}

\usepackage{listings}
\usepackage{pifont}
\usepackage{color}
\usepackage{bm}
\usepackage{array}

\DeclareMathOperator*{\argmax}{argmax}

\newcommand{\ba}{\mathbf{a}}
\newcommand{\bb}{\mathbf{b}}
\newcommand{\bc}{\mathbf{c}}
\newcommand{\br}{\mathbf{r}}
\newcommand{\bw}{\mathbf{w}}

\newcommand{\bmu}{\bm{\mu}}
\newcommand{\bsi}{\bm{\Sigma}}
\newcommand{\ins}{s_{\textbf{I}}}
\newcommand{\outs}{s_{\textbf{O}}}

\newcommand{\Cn}{\mathcal{N}}
\newcommand{\Cl}{\mathcal{L}}

\newcommand{\Ra}{\shortrightarrow}
\newcommand{\iprod}[1]{\left\langle{#1}\right\rangle}

\aclfinalcopy 

\title{Gaussian Mixture Latent Vector Grammars}

\author{Yanpeng Zhao, Liwen Zhang, Kewei Tu\\
	School of Information Science and Technology,\\
	ShanghaiTech University, Shanghai, China \\
	{\tt \{zhaoyp1,zhanglw1,tukw\}@shanghaitech.edu.cn}}

\date{}

\begin{document}
	\maketitle
	\begin{abstract}
		We introduce Latent Vector Grammars (LVeGs), a new framework that extends latent variable grammars such that each nonterminal symbol is associated with a continuous vector space representing the set of (infinitely many) subtypes of the nonterminal.
		We show that previous models such as latent variable grammars and compositional vector grammars can be interpreted as special cases of LVeGs.
		We then present Gaussian Mixture LVeGs (GM-LVeGs), a new special case of LVeGs that uses Gaussian mixtures to formulate the weights of production rules over subtypes of nonterminals.
		A major advantage of using Gaussian mixtures is that the partition function and the expectations of subtype rules can be computed using an extension of the inside-outside algorithm,
		which enables efficient inference and learning.
		We apply GM-LVeGs to part-of-speech tagging and constituency parsing and show that GM-LVeGs can achieve competitive accuracies.
		Our code is available at \href{https://github.com/zhaoyanpeng/lveg}{https://github.com/zhaoyanpeng/lveg}.
	\end{abstract}
	
	\section{Introduction}
	In constituency parsing, refining coarse syntactic categories of treebank grammars~\citep{charniak1996tree} into fine-grained subtypes has been proven effective in improving parsing results.
	Previous approaches to refining syntactic categories use tree annotations~\citep{johnson1998pcfg}, lexicalization~\citep{charniak2000maximum,collins2003head}, or linguistically motivated category splitting~\citep{klein2003accurate}.
	\citet{matsuzaki2005probabilistic} introduce latent variable grammars,
	in which each syntactic category (represented by a nonterminal) is split into a fixed number of subtypes
	and a discrete latent variable is used to indicate the subtype of the nonterminal when it appears in a specific parse tree.
	Since the latent variables are not observable in treebanks, 
	the grammar is learned using expectation-maximization. 
	\citet{petrov2006learning} present a split-merge approach to learning latent variable grammars,
	which hierarchically splits each nonterminal and merges ineffective splits.
	\citet{petrov2008sparse} further allow a nonterminal to have different splits in different production rules, 
	which results in a more compact grammar.


	Recently, neural approaches become very popular in natural language processing (NLP).
	An important technique in neural approaches to NLP is to represent discrete symbols such as words and syntactic categories with continuous vectors or embeddings.
	Since the distances between such vector representations often reflect the similarity between the corresponding symbols,
	this technique facilitates more informed smoothing in learning functions of symbols (e.g., the probability of a production rule).
	In addition, what a symbol represents may subtly depend on its context,
	and a continuous vector representation has the potential of representing each instance of the symbol in a more precise manner.
	For constituency parsing, 
	recursive neural networks~\cite{socher2011parsing} and their extensions such as compositional vector grammars~\cite{socher2013parsing} can be seen as representing nonterminals in a context-free grammar with continuous vectors.
	However, 
	exact inference in these models is intractable.

	In this paper, we introduce latent vector grammars (LVeGs), a novel framework of grammars with fine-grained nonterminal subtypes.
	A LVeG associates each nonterminal with a continuous vector space that represents the set of (infinitely many) subtypes of the nonterminal.
	For each instance of a nonterminal that appears in a parse tree, 
	its subtype is represented by a latent vector.
	For each production rule over nonterminals,
	a non-negative continuous function specifies the weight of any fine-grained production rule over subtypes of the nonterminals.
	Compared with latent variable grammars which assume a small fixed number of subtypes for each nonterminal,
	LVeGs assume an unlimited number of subtypes and are potentially more expressive.
	By having weight functions of varying smoothness for different production rules,
	LVeGs can also control the level of subtype granularity for different productions,
	which has been shown to improve the parsing accuracy~\cite{petrov2008sparse}.
	In addition, similarity between subtypes of a nonterminal can be naturally modeled by the distance between the corresponding vectors,
	so by using continuous and smooth weight functions we can ensure that similar subtypes will have similar syntactic behaviors.

	We further present Gaussian Mixture LVeGs (GM-LVeGs), a special case of LVeGs that uses mixtures of Gaussian distributions as the weight functions of fine-grained production rules.
	A major advantage of GM-LVeGs is that the partition function and the expectations of fine-grained production rules can be computed using an extension of the inside-outside algorithm.
	This makes it possible to efficiently compute the gradients during discriminative learning of GM-LVeGs.
	We evaluate GM-LVeGs on part-of-speech tagging and constituency parsing on a variety of languages and corpora and show that GM-LVeGs achieve competitive results.

	It shall be noted that many modern state-of-the-art constituency parsers predict how likely a constituent is based on not only local information (such as the production rules used in composing the constituent),
	but also contextual information of the constituent.
	For example, the neural CRF parser~\cite{durrett2015neural} looks at the words before and after the constituent; and RNNG~\cite{dyer2016recurrent} looks at the constituents that are already predicted (in the stack) and the words that are not processed (in the buffer).
	In this paper, however, we choose to focus on the basic framework and algorithms of LVeGs and leave the incorporation of contextual information for future work.
	We believe that by laying a solid foundation for LVeGs,
	our work can pave the way for many interesting extensions of LVeGs in the future.

	\section{Latent Vector Grammars}
	A latent vector grammar (LVeG) considers subtypes of nonterminals as continuous vectors
	and associates each nonterminal with a latent vector space representing the set of its subtypes.
	For each production rule, 
	the LVeG defines a weight function over the subtypes of the nonterminal involved in the production rule.
	In this way, it models the space of refinements of the production rule.

	\subsection{Model Definition}
	A latent vector grammar is defined as a 5-tuple $\mathcal{G} = (N, S, \Sigma, R, W)$, 
	where $N$ is a finite set of nonterminal symbols, $S\in N$ is the start symbol, 
	$\Sigma$ is a finite set of terminal symbols such that $N \cap \Sigma = \varnothing$,
	$R$ is a set production rules of the form $X\Ra \gamma$ where $X\in N$ and $\gamma \in (N\cup \Sigma)^*$,
	$W$ is a set of rule weight functions 
	indexed by production rules in $R$ (to be defined below).
	In the following discussion,
	we consider $R$ in the Chomsky normal form (CNF) for clarity of presentation.
	However, it is straightforward to extend our formulation to the general case.

	Unless otherwise specified, 
	we always use capital letters $A, B, C,\ldots$ for nonterminal symbols  
	and use bold lowercase letters $\ba, \bb, \bc,\ldots$ for their subtypes. 
	Note that subtypes are represented by continuous vectors.
	For a production rule of the form $A\Ra BC$, 
	its weight function is $W_{A\Ra BC}(\ba, \bb, \bc)$.
	For a production rule of the form $A\Ra w$ where $w\in \Sigma$,
	its weight function is $W_{A\Ra w}(\ba)$.
	The weight functions should be non-negative, continuous and smooth, 
	and hence fine-grained production rules of similar subtypes of a nonterminal would have similar weight assignments.
	Rule weights can be normalized such that 
	$\sum_{B,C} \int_{\bb,\bc} W_{A\Ra BC}(\ba,\bb,\bc) d\bb d\bc = 1$,
	which leads to a probabilistic context-free grammar (PCFG). 
	Whether the weights are normalized or not leads to different model classes and accordingly different estimation methods.
	However, the two model classes are proven equivalent by~\citet{smith2007weighted}.

	\subsection{Relation to Other Models}
	Latent variable grammars (LVGs)~\citep{matsuzaki2005probabilistic, petrov2006learning} associate each nonterminal with a discrete latent variable,
	which is used to indicate the subtype of the nonterminal when it appears in a parse tree.
	Through nonterminal-splitting and the expectation-maximization algorithm, 
	fine-grained production rules can be automatically induced from a treebank.

	We show that LVGs can be seen as a special case of LVeGs. 
	Specifically, we can use one-hot vectors in LVeGs to represent latent variables in LVGs
	and define weight functions in LVeGs accordingly.
	Consider a production rule $r: A\Ra BC$.
	In a LVG, each nonterminal is split into a number of subtypes. 
	Suppose $A$, $B$, and $C$ are split into $n_A$, $n_B$, and $n_C$ subtypes respectively. 
	$a_x$ is the $x$-th subtype of $A$, 
	$b_y$ is the $y$-th subtype of $B$, 
	and $c_z$ is the $z$-th subtype of $C$.
	$a_x\Ra b_y c_z$ is a fine-grained production rule of $A\Ra BC$,
	where $x=1,\ldots, n_A$, $y=1,\ldots, n_B$, and $z=1,\ldots, n_C$. 
	The probabilities of all the fine-grained production rules can be represented by a rank-3 tensor $\Theta_{A\Ra BC} \in \mathbb{R}^{n_{A} \times n_{B} \times n_{C}}$. 
	To cast the LVG as a LVeG, we require that the latent vectors in the LVeG must be one-hot vectors.
	We achieve this by defining weight functions that output zero if any of the input vectors is not one-hot.
	Specifically, we define the weight function of the production rule $A\Ra BC$ as:
	\begin{align}\label{eq:weight}
	W_{r}(\ba, \bb, \bc) &= \sum_{x, y, z}\Theta_{A\Ra BC}\bc\bb\ba \times \left(\delta(\ba-\ba_{x}) \right. \nonumber\\
	&\times \left. \delta(\bb-\bb_{y})\times\delta(\bc-\bc_{z}) \right)\,,
	\end{align}
	where $\delta(\cdot)$ is the Dirac delta function, $\ba_x\in \mathbb{R}^{n_A}$, $\bb_y\in \mathbb{R}^{n_B}$, $\bc_z\in \mathbb{R}^{n_C}$ are one-hot vectors 
	(which are zero everywhere with the exception of a single 1 at the $x$-th index of $\ba_x$, the $y$-th index of $\bb_y$, and the $z$-th index of $\bc_z$)
	and $\Theta_{A\Ra BC}$ is multiplied sequentially by $\bc$, $\bb$, and $\ba$.
	
	
	Compared with LVGs, LVeGs have the following advantages.
	While a LVG contains a finite, typically small number of subtypes for each nonterminal,
	a LVeG uses a continuous space to represent an infinite number of subtypes.
	When equipped with weight functions of sufficient complexity,
	LVeGs can represent more fine-grained syntactic categories and production rules than LVGs.
	By controlling the complexity and smoothness of the weight functions,
	a LVeG is also capable of representing any level of subtype granularity.
	Importantly, this allows us to change the level of subtype granularity for the same nonterminal in different production rules,
	which is similar to multi-scale grammars~\citep{petrov2008sparse}.
	In addition, with a continuous space of subtypes in a LVeG,
	similarity between subtypes can be naturally modeled by their distance in the space
	and can be automatically learned from data.
	Consequently, with continuous and smooth weight functions,
	fine-grained production rules over similar subtypes would have similar weights in LVeGs, 
	eliminating the need for the extra smoothing steps that are necessary in training LVGs.

	Compositional vector grammars (CVGs) ~\citep{socher2013parsing},
	an extension of recursive neural networks (RNNs)~\cite{socher2011parsing},
	can also be seen as a special case of LVeGs.
	For a production rule $r: A\Ra BC$,
	a CVG can be interpreted as specifying its weight function $W_{r}(\ba, \bb, \bc)$ in the following way.
	First, a neural network $f$ indexed by $B$ and $C$ is used to compute a parent vector
	$\mathbf{p} = f_{BC}(\bb, \bc)$.
	Next, the score of the parent vector is computed using a base PCFG and a vector
	$\mathbf{v}_{BC}$:
	\begin{eqnarray}
	s(\mathbf{p}) = \mathbf{v}_{BC}^{T}\mathbf{p} + \log P(A\Ra BC)\,,
	\end{eqnarray}
	where $P(A\Ra BC)$ is the rule probability from the base PCFG.
	Then, the weight function of the production rule $A\Ra BC$ is defined as:
	\begin{eqnarray}
	W_{r}(\ba, \bb, \bc) = \exp\left(s(\mathbf{p})\right) \times \delta(\ba - \mathbf{p})\,.
	\end{eqnarray}

	This form of weight functions in CVGs leads to point estimation of latent vectors in a parse tree,
	i.e., for each nonterminal in a given parse tree,
	only one subtype in the whole subtype space would lead to a non-zero weight of the parse.
	In addition, different parse trees of the same substring typically lead to different point estimations of the subtype vector at the root nonterminal.
	Consequently, CVGs cannot use dynamic programming for inference and hence have to resort to greedy search or beam search.

	\section{Gaussian Mixture LVeGs}
	A major challenge in applying LVeGs to parsing is that it is impossible to enumerate the infinite number of subtypes.
	Previous work such as CVGs resorts to point estimation and greedy search.
	In this section we present Gaussian Mixture LVeGs (GM-LVeGs),
	which use mixtures of Gaussian distributions as the weight functions in LVeGs.
	Because Gaussian mixtures have the nice property of being closed under product, summation, and marginalization,
	we can compute the partition function and the expectations of fine-grained production rules using dynamic programming.
	This in turn makes efficient learning and parsing possible.

	\subsection{Representation}
	In a GM-LVeG, the weight function of a production rule $r$ is defined as a Gaussian mixture containing $K_r$ mixture components:
	\begin{eqnarray}
	W_{r}(\br) = \sum_{k = 1}^{K_{r}} \rho_{r, k}\, \Cn(\br| \bmu_{r, k},\bsi_{r, k})\,,
	\end{eqnarray}
	where $\br$ is the concatenation of the latent vectors of the nonterminals in $r$, which denotes a fine-grained production rule of $r$.
	$\rho_{r, k} > 0$ is the $k$-th mixture weight (the mixture weights do not necessarily sum up to 1),
	$\Cn(\br | \bmu_{r, k}, \bsi_{r, k})$ is the $k$-th Gaussian distribution parameterized by mean $\bmu_{r, k}$ and covariance matrix $\bsi_{r, k}$,
	and $K_r$ is the number of mixture components, which can be different for different production rules.
	Below we write $\Cn(\br| \bmu_{r, k},\bsi_{r, k})$ as $\Cn_{r,k}(\br)$ for brevity.
	Given a production rule of the form $A\Ra BC$, 
	the GM-LVeG expects $\br = \left[\ba; \bb; \bc\right]$ and $\ba, \bb, \bc \in \mathbb{R}^{d}$,
	where $d$ is the dimension of the vectors $\ba, \bb, \bc$.
	We use the same dimension for all the subtype vectors.

	For the sake of computational efficiency, we use diagonal or spherical Gaussian distributions, whose covariance matrices are diagonal,
	so that the inverse of covariance matrices in Equation~\ref{eq:grad_dmu}--\ref{eq:grad_dsig} can be computed in linear time.
	A spherical Gaussian has a diagonal covariance matrix where all the diagonal elements are equal, so it has fewer free parameters than a diagonal Gaussian and results in faster learning and parsing.
	We empirically find that spherical Gaussians lead to slightly better balance between the efficiency and the parsing accuracy than diagonal Gaussians.

	\subsection{Parsing}\label{sec:parsing}
	The goal of parsing is to find the most probable parse tree $T^*$ with unrefined nonterminals for a sentence $\bw$ of $n$ words $w_{1:n} = w_1\ldots w_n$. 
	This is formally defined as: 
	\begin{eqnarray}
	\label{eq:parse_obj}
	T^* = \argmax_{T\in G(\bw)} P(T |\bw)\,,
	\end{eqnarray}
	where $G(\bw)$ denotes the set of parse trees with unrefined nonterminals for $\bw$.
	In a PCFG, $T^*$ can be found using dynamic programming such as the CYK algorithm.
	However, parsing becomes intractable with LVeGs,
	and even with LVGs, the special case of LVeGs.

	\begin{table*}[!ht]
		\centering
		{\setlength{\tabcolsep}{.0em}
			\begin{tabular}{c}
				\toprule 
				\begin{minipage}{\linewidth}
					\vspace{-.5em}
					\begin{eqnarray}
					\label{eq:inside}
					\ins^{A}(\ba, i, j) =& \underset{A\Ra BC \in R}{\sum} \,\, \underset{k=i,\cdots,j-1}{\sum} & \iint W_{A\Ra BC}(\ba,\bb,\bc) \times \ins^{B}(\bb, i,k) \times \ins^{C}(\bc, k + 1,j) \,d\bb d\bc\,. \\ 
					\label{eq:outside} 
					\outs^{A}(\ba, i, j) =& \underset{B\Ra CA \in R}{\sum} \,\, \underset{k=1,\cdots,i-1}{\sum} & \iint W_{B\Ra CA}(\bb,\bc,\ba) \times \outs^{B}(\bb, k, j) \times \ins^{C}(\bc, k, i - 1) \,d\bb d\bc \nonumber \\
					+& \underset{B\Ra AC \in R}{\sum} \,\, \underset{k=j+1,\cdots,n}{\sum} & \iint W_{B\Ra AC}(\bb,\ba,\bc) \times \outs^{B}(\bb, i, k) \times \ins^{C}(\bc, j + 1, k) \,d\bb d\bc\,.
					\end{eqnarray}
					\vspace{-.5em}
				\end{minipage}\\
				\toprule
				\begin{minipage}{\linewidth}
					\vspace{-.5em}
					\begin{eqnarray}
					\label{eq:rscore}
					s(A\Ra BC, i, k , j) =\! \iiint W_{A\Ra BC}(\ba,\bb,\bc) \times \outs^{A}(\ba, i, j) \times \ins^{B}(\bb, i, k) \times \ins^{C}(\bc, k + 1, j) \,d\ba d\bb d\bc\,. \!\!\!\!\!
					\end{eqnarray}
					\vspace{-.5em}
				\end{minipage}\\
				\bottomrule
		\end{tabular}}
		\caption{
			\label{tab:eqs}
			Equation~\ref{eq:inside}:
			$\ins^{A}(\ba, i, j)$ is the inside score of a nonterminal $A$ over a span $w_{i:j}$ in the sentence $w_{1:n}$, where $1 \le i < j \le n$.
			Equation~\ref{eq:outside}:
			$\outs^{A}(\ba, i, j)$ is the outside score of a nonterminal $A$ over a span $w_{i:j}$ in the sentence $w_{1:n}$, where $1 \le i \le j \le n$.
			Equation~\ref{eq:rscore}:  
			$s(A\Ra BC, i, k, j)$ is the score of a production rule $A\Ra BC$ with nonterminals $A$, $B$, and $C$ spanning words $w_{i:j}$, $w_{i, k}$, and $w_{k + 1:j}$ respectively in the sentence $w_{1:n}$,
			where $1 \le i \le k < j \le n$.}
	\end{table*}

	A common practice in parsing with LVGs is to use max-rule parsing~\cite{petrov2006learning,petrov2007improved}.
	The basic idea of max-rule parsing is to decompose the posteriors over parses into the posteriors over production rules approximately.
	This requires calculating the expected counts of unrefined production rules in parsing the input sentence.
	Since Gaussian mixtures are closed under product, summation, and marginalization,
	in GM-LVeGs the expected counts can be calculated using the inside-outside algorithm in the following way.
	Given a sentence $w_{1:n}$,
	we first calculate the inside score $\ins^{A}(\ba, i, j)$ and outside score $\outs^{A}(\ba, i, j)$ 
	for a nonterminal $A$ over a span $w_{i:j}$ using Equation~\ref{eq:inside} and Equation~\ref{eq:outside} in Table~\ref{tab:eqs} respectively.
	Note that both $\ins^{A}(\ba, i, j)$ and $\outs^{A}(\ba, i, j)$ are mixtures of Gaussian distributions of the subtype vector $\ba$.
	Next, using Equation~\ref{eq:rscore} in Table~\ref{tab:eqs}, 
	we calculate the score $s(A\Ra BC, i, k, j)$ ($1 \le i \le k < j \le n$),
	where $\iprod{A\Ra BC, i, k, j}$ represents a production rule $A\Ra BC$ with nonterminals $A$, $B$, and $C$ spanning words $w_{i:j}$, $w_{i, k}$, and $w_{k + 1:j}$ respectively in the sentence $w_{1:n}$.
	Then the expected count (or posterior) of $\iprod{A\Ra BC, i, k, j}$ is calculated as:
	\begin{eqnarray}
	q(A\Ra BC, i, k, j) = \frac{s(A\Ra BC, i, k, j)}{\ins(S, 1, n)}\,,
	\end{eqnarray}
	where $\ins(S, 1, n)$ is the inside score for the start symbol $S$ spanning the whole sentence $w_{1:n}$.
	After calculating all the expected counts, 
	we can use the \textsc{Max-Rule-Product} algorithm~\citep{petrov2007improved} for parsing,
	which returns a parse with the highest probability that all the production rules are correct.
	Its objective function is given by
	\begin{eqnarray}
	T_q^{*} = \argmax_{T\in G(\bw)} \prod_{e \in T} q(e)\,,
	\end{eqnarray}
	where $e$ ranges over all the 4-tuples $\iprod{A\Ra BC, i, k, j}$ in the parse tree $T$.
	This objective function can be efficiently solved by dynamic programming such as the CYK algorithm.

	Although the time complexity of the inside-outside algorithm with GM-LVeGs is polynomial in the sentence length and the nonterminal number,
	in practice the algorithm is still slow
	because the number of Gaussian components in the inside and outside scores increases dramatically with the recursion depth. 
	To speed up the computation,
	we prune Gaussian components in the inside and outside scores using the following technique.
	Suppose we have a minimum pruning threshold $k_{min}$ and a maximum pruning threshold $k_{max}$.
	Given an inside or outside score with $k_{c}$ Gaussian components,
	if $k_{c} \le k_{min}$,
	then we do not prune any Gaussian component;
	otherwise, we compute $k_{allow} = \min\{k_{min} + \text{floor}(k_{c}^{\vartheta}), k_{max}\}$ ($0\le\vartheta\le1$ is a constant)
	and keep only $k_{allow}$ components with the largest mixture weights.

	In addition to component pruning,
	we also employ two constituent pruning techniques to reduce the search space during parsing.
	The first technique is used by~\citet{petrov2006learning}.
	Before parsing a sentence with a GM-LVeG,
	we run the inside-outside algorithm with the treebank grammar
	and calculate the posterior probability of every nonterminal spanning every substring.
	Then a nonterminal would be pruned from a span if its posterior probability is below a pre-specified threshold $p_{min}$.
	When parsing with GM-LVeGs, we only consider the unpruned nonterminals for each span.

	The second constituent pruning technique is similar to the one used by~\citet{socher2013parsing}.
	Note that for a strong constituency parser such as the Berkeley parser~\citep{petrov2007improved},
	the constituents in the top 200 best parses of a sentence can cover almost all the constituents in the gold parse tree.
	So we first use an existing constituency parser to run $k$-best parsing with $k = 200$ on the input sentence.
	Then we parse with a GM-LVeG and only consider the constituents that appear in the top 200 parses.
	Note that this method is different from the re-ranking technique 
	because it may produce a parse different from the top 200 parses.

	\subsection{Learning}
	Given a training dataset $D = \{(T_i, \bw_i)\,|\, i = 1,\ldots,m\}$ containing $m$ samples, 
	where $T_i$ is the gold parse tree with unrefined nonterminals for the sentence $\bw_i$, the objective of discriminative learning is to minimize the negative log conditional likelihood:
	\begin{eqnarray}
	\label{eq:learn_obj}
	\Cl(\Theta) = -\log \prod_{i = 1}^{m} P(T_{i} | \bw_i; \Theta)\,,
	\end{eqnarray}
	where $\Theta$ represents the set of parameters of the GM-LVeG.

	We optimize the objective function using 
	the Adam~\citep{kingma2014adam} optimization algorithm. The derivative with respect to $\Theta_r$, 
	the parameters of the weight function $W_r(\br)$ of an unrefined production rule $r$, is calculated as follows 
	(the derivation is in the supplementary material):
	\begin{align}
	\label{eq:grad}
	\frac{\partial\Cl(\Theta)}{\partial\Theta_r} \! =& \sum_{i=1}^{m} \int
	\left( 
	\frac{\partial W_r(\br)} {\partial\Theta_r}
	\right. \\
	\! \times& \left.
	\frac{\mathbb{E}_{P(t|\bw_i)}[f_r(t)] - \mathbb{E}_{P(t|T_i)}[f_r(t)]}{W_r(\br)} 
	\right)\,d\br\,, \nonumber
	\end{align}
	where $t$ indicates a parse tree with nonterminal subtypes,
	and $f_r(t)$ is the number of occurrences of the unrefined rule $r$ in the unrefined parse tree
	that is obtained by replacing all the subtypes in $t$ with the corresponding nonterminals.
	The two expectations in Equation~\ref{eq:grad} can be efficiently computed using the inside-outside algorithm.
	Because the second expectation is conditioned on the parse tree $T_i$,
	in Equation~\ref{eq:inside} and Equation~\ref{eq:outside} we can skip all the summations and assign the values of $B$, $C$, and $k$ according to $T_i$.

	In GM-LVeGs, $\Theta_r$ is the set of parameters in a Gaussian mixture:
	\begin{eqnarray}
	\Theta_r = \{(\rho_{r, k}, \bmu_{r, k}, \bsi_{r, k}) | k = 1,\ldots,K_r\}\,.
	\end{eqnarray}
	According to Equation~\ref{eq:grad}, we need to take the derivatives of $W_r(\br)$ respect to $\rho_{r, k}$, $\bmu_{r, k}$, and $\bsi_{r, k}$ respectively:
	\begin{align}
	\label{eq:grad_dmw}
	\!\!\!\!\partial W_r(\br) / \partial\rho_{r, k} &= \Cn_{r, k}(\br) \,, \\
	\label{eq:grad_dmu}
	\!\!\!\!\partial W_r(\br) / \partial\bmu_{r, k} &= \rho_{r, k}\Cn_{r, k}(\br)  \bsi_{r, k}^{-1}  (\br - \bmu_{r, k}) \,,\!\!\!\! \\
	\label{eq:grad_dsig}
	\!\!\!\!\partial W_r(\br) / \partial\bsi_{r, k} &= \rho_{r, k}\Cn_{r, k}(\br)  \bsi_{r, k}^{-1} \frac{1}{2} \left( \vphantom{ \bsi_{r, k}^{-1} } -I \right. \\
	&+ \left. (\br - \bmu_{r, k}) (\br - \bmu_{r, k})^{T} \bsi_{r, k}^{-1} \right)\,. \nonumber
	\end{align}
	Substituting Equation~\ref{eq:grad_dmw}--\ref{eq:grad_dsig} into Equation~\ref{eq:grad}, we have the full gradient formulations of all the parameters.
	In spite of the integral in Equation~\ref{eq:grad}, we can derive a closed-form solution for the gradient of each parameter, which is shown in the supplementary material.
	
	
	In order to keep each mixture weight $\rho_{r, k}$ positive,
	we do not directly optimize $\rho_{r, k}$;
	instead, we set 
	$\rho_{r, k} = \exp (\theta_{\rho_{r, k}})$
	and optimize $\theta_{\rho_{r, k}}$ by gradient descent.
	We use a similar trick to keep each covariance matrix $\bsi_{r, k}$ positive definite.
	
	Since we use the inside-outside algorithm described in Section~\ref{sec:parsing} to calculate the two expectations in Equation~\ref{eq:grad},
	we face the same efficiency problem that we encounter in parsing.
	To speed up the computation,
	we again use both component pruning and constituent pruning introduced in Section~\ref{sec:parsing}.

	Because gradient descent is often sensitive to the initial values of the parameters, 
	we employ the following informed initialization method.
	Mixture weights are initialized using the treebank grammar. 
	Suppose in the treebank grammar $P(r)$ is the probability of a production rule $r$.
	We initialize the mixture weights in the weight function $W_r$ by 
	$\rho_{r, k} = \alpha \cdot P(r)$ where 
	$\alpha > 1$ is a constant. 
	We initialize all the covariance matrices to identity matrices 
	and initialize each mean with a value uniformly sampled from $[-0.05, 0.05]$.


	\section{Experiment}
	We evaluate the GM-LVeG on part-of-speech (POS) tagging and constituency parsing
	and compare it against its special cases such as LVGs and CVGs.
	It shall be noted that in this paper we focus on the basic framework of LVeGs and aim to show its potential advantage over previous special cases. 
	It is therefore not our goal to compete with the latest state-of-the-art approaches to tagging and parsing. 
	In particular, we currently do not incorporate contextual information of words and constituents during tagging and parsing, 
	while such information is critical in achieving state-of-the-art accuracy.
	We will discuss future improvements of LVeGs in Section~\ref{sec:dis}.
	
	\subsection{Datasets}
	\textbf{Parsing.}
	We use the Wall Street Journal corpus from the Penn English Treebank (WSJ)~\citep{marcus1994penn}. 
	Following the standard data splitting,
	we use sections 2 to 21 for training, section 23 for testing, and section 22 for development.
	We pre-process the treebank using a right-branching binarization procedure to obtain an unannotated X-bar grammar,
	so that there are only binary and unary production rules.
	To deal with the problem of unknown words in testing,
	we adopt the unknown word features used in the Berkeley parser
	and set the unknown word threshold to 1.
	Specifically, any word occurring less than two times is replaced by one of the 60 unknown word categories.
	
	\noindent\textbf{Tagging.}
	(1) We use Wall Street Journal corpus from the Penn English Treebank (WSJ)~\citep{marcus1994penn}. 
	Following the standard data splitting,
	we use sections 0 to 18 for training, sections 22 to 24 for testing, and sections 19 to 21 for development.
	(2) The Universal Dependencies treebank 1.4 (UD)~\citep{nivre2016universal}, in which English, French, German, Russian, Spanish,  Indonesian,  Finnish, and Italian treebanks are used. 
	We use the original data splitting of these corpora for training and testing.
	For both WSJ and UD English treebanks, we deal with unknown words in the same way as we do in parsing. 
	For the rest of the data, we use only one unknown word category
	and the unknown word threshold is also set to 1.
	
	
	\subsection{POS Tagging}
	POS tagging is the task of labeling each word in a sentence with the most probable part-of-speech tag.
	Here we focus on POS tagging with Hidden Markov Models (HMMs).
	Because HMMs are equivalent to probabilistic regular grammars, 
	we can extend HMMs with both LVGs and LVeGs.
	Specifically, the hidden states in HMMs can be seen as nonterminals in regular grammars
	and therefore can be associated with latent variables or latent vectors.

	We implement two training methods for LVGs.
	The first (LVG-G) is generative training using expectation-maximization that maximizes the joint probability of the sentence and the tags.
	The second (LVG-D) is discriminative training using gradient descent that maximizes the conditional probability of the tags given the sentence.
	In both cases, each nonterminal is split into a fixed number of subtypes.
	In our experiments we test 1, 2, 4, 8, and 16 subtypes of each nonterminal.
	Due to the limited space, we only report experimental results of LVG with 16 subtypes for each nonterminal.
	Full experimental results can be found in the supplementary material.

	We experiment with two different GM-LVeGs:
	GM-LVeG-D with diagonal Gaussians
	and GM-LVeG-S with spherical Gaussians.
	In both cases,
	we fix the number of Gaussian components $K_r$ to 4 and the dimension of the latent vectors $d$ to 3.
	We do not use any pruning techniques in learning and inference
	because we find that our algorithm is fast enough with the current setting of $K_r$ and $d$.
	We train the GM-LVeGs for 20 epoches and select the models with the best token accuracy on the development data for the final testing.
	
	We report both token accuracy and sentence accuracy of POS tagging in Table~\ref{tab:tag_acc}. 
	It can be seen that, on all the testing data, 
	GM-LVeGs consistently surpass LVGs in terms of both token accuracy and sentence accuracy.
	GM-LVeG-D is slightly better than GM-LVeG-S in sentence accuracy,
	producing the best sentence accuracy on 5 of the 9 testing datasets.
	GM-LVeG-S performs slightly better than GM-LVeG-D in token accuracy on 5 of the 9 datasets.
	Overall, there is not significant difference between GM-LVeG-D and GM-LVeG-S.
	However, GM-LVeG-S admits more efficient learning than GM-LVeG-D in practice
	since it has fewer parameters.

	\begin{table*}[!ht]\small
		\centering
		\vskip -.0in
		{\setlength{\tabcolsep}{.3em}
			\makebox[\linewidth]{\resizebox{\linewidth}{!}{%
					\begin{tabular}{lcccccccccccccccccc}
						\toprule
						\multirow{3}{*}{Model} &
						\multicolumn{2}{c}{WSJ}  &
						\multicolumn{2}{c}{English}  &
						\multicolumn{2}{c}{French} &
						\multicolumn{2}{c}{German} &
						\multicolumn{2}{c}{Russian} &
						\multicolumn{2}{c}{Spanish} &
						\multicolumn{2}{c}{Indonesian} &
						\multicolumn{2}{c}{Finnish} &
						\multicolumn{2}{c}{Italian} \\ 
						\cmidrule(lr){2-3} \cmidrule(lr){4-5} \cmidrule(lr){6-7} \cmidrule(lr){8-9} \cmidrule(lr){10-11} \cmidrule(lr){12-13} \cmidrule(lr){14-15} \cmidrule(lr){16-17} \cmidrule(lr){18-19} 
						&
						\multicolumn{1}{c}{T}  & \multicolumn{1}{c}{S} &
						\multicolumn{1}{c}{T}  & \multicolumn{1}{c}{S} &
						\multicolumn{1}{c}{T}  & \multicolumn{1}{c}{S} &
						\multicolumn{1}{c}{T}  & \multicolumn{1}{c}{S} &
						\multicolumn{1}{c}{T}  & \multicolumn{1}{c}{S} &
						\multicolumn{1}{c}{T}  & \multicolumn{1}{c}{S} &
						\multicolumn{1}{c}{T}  & \multicolumn{1}{c}{S} &
						\multicolumn{1}{c}{T}  & \multicolumn{1}{c}{S} &
						\multicolumn{1}{c}{T}  & \multicolumn{1}{c}{S} \\
						\midrule
						LVG-D-16& 96.62& 48.74& 92.31& 52.67& 93.75& 34.90& 87.38& 20.98& 81.91& 12.25& 92.47& 24.82& 89.27& 20.29& 83.81& 19.29& 94.81& 45.19\\
						LVG-G-16& 96.78& 50.88& 93.30& 57.54& 94.52& 34.90& 88.92& 24.05& 84.03& 16.63& 93.21& 27.37& 90.09& 21.19& 85.01& 20.53& 95.46& 48.26\\
						\toprule
						GM-LVeG-D& 96.99& 53.10& \textbf{93.66}& \textbf{59.46}& 94.73& \textbf{39.60}& 89.11& 24.77& \textbf{84.21}& 17.84& \textbf{93.76}& \textbf{32.48}& \textbf{90.24}& \textbf{21.72}& 85.27& \textbf{23.30}& 95.61& \textbf{50.72}\\
						GM-LVeG-S& \textbf{97.00}& \textbf{53.11}& 93.55& 58.11& \textbf{94.74}& 39.26& \textbf{89.14}& \textbf{25.58}& 84.06& \textbf{18.44}& 93.52& 30.66& 90.12& \textbf{21.72}& \textbf{85.35}& 22.07& \textbf{95.62}& 49.69\\
						\bottomrule
		\end{tabular}}}}
		\caption{\label{tab:tag_acc}Token accuracy (T) and sentence accuracy (S) for POS tagging on the testing data. }
		\vskip -.0in
	\end{table*}

	\subsection{Parsing}
	For efficiency, we train GM-LVeGs only on sentences with no more than 50 words (totally 39115 sentences).
	Since we have found that spherical Gaussians are better than diagonal Gaussians considering both model performance and learning efficiency,
	here we use spherical Gaussians in the weight functions.
	The dimension of latent vectors $d$ is set to 3,
	and all the Gaussian mixtures have $K_r = 4$ components.
	We use $\alpha = 8$ in initializing mixture weights.
	We train the GM-LVeG for 15 epoches 
	and select the model with the highest F1 score on the development data for the final testing. 
	We use component pruning in both learning and parsing, 
	with $k_{max} = 50$ and $\vartheta = 0.35$ in both learning and parsing,
	$k_{min} = 40$ in learning and $k_{min} = 20$ in parsing.
	During learning we use the first constituent pruning technique with the pruning threshold $p_{min} = 1e-5$,
	and during parsing we use the second constituent pruning technique based on the Berkeley parser which produced 133 parses on average for each testing sentence.
	As can be seen, we use weaker pruning during training than during testing. 
	This is because in training stronger pruning (even if accurate) results in worse estimation of the first expectation in Equation~\ref{eq:grad}, which makes gradient computation less accurate.

	We compare LVeGs with CVGs and several variants of LVGs:
	(1) LVG-G-16 and LVG-D-16, which are LVGs with 16 subtypes for each nonterminal with discriminative and generative training respectively (accuracies obtained from~\citet{petrov2008discriminative});
	(2) Multi-scale grammars~\citep{petrov2008sparse}, trained without using the span features in order for a fair comparison;
	(3) Berkeley parser~\citep{petrov2007improved} (accuracies obtained from~\citet{petrov2008sparse} because~\citet{petrov2007improved} do not report exact match scores).
	The experimental results are shown in Table~\ref{tab:parse_acc}.
	It can be seen that GM-LVeG-S produces the best F1 scores on both the development data and the testing data.
	It surpasses the Berkeley parser by 0.92\% in F1 score on the testing data.
	Its exact match score on the testing data is only slightly lower than that of LVG-D-16.

	\begin{table}[!ht]\small
		\centering
		\vskip .055in
		{\setlength{\tabcolsep}{.5em}
			\makebox[\linewidth]{\resizebox{\linewidth}{!}{%
					\begin{tabular}{ccccccc}
						\toprule
						\multirow{3}{*}{Model} & 
						\multicolumn{1}{c}{dev (all)} &
						\multicolumn{2}{c}{$\text{test} \le 40$} & 
						\multicolumn{2}{c}{test (all)} \\
						\cmidrule(lr){2-2} \cmidrule(lr){3-4} \cmidrule(lr){5-6}
						&
						\multicolumn{1}{c}{F1} & 
						\multicolumn{1}{c}{F1} & \multicolumn{1}{c}{EX} &
						\multicolumn{1}{c}{F1}  & \multicolumn{1}{c}{EX} \\
						\toprule
						LVG-G-16 &  & &  & 88.70 & 35.80 &\\
						LVG-D-16 &  & & & 89.30 & \textbf{39.40} \\
						Multi-Scale & & 89.70 & 39.60 & 89.20 & 37.20 \\
						Berkeley Parser & & 90.60 & 39.10 & 90.10 & 37.10 \\
						CVG (SU-RNN) & 91.20 & 91.10 & & 90.40 & \\
						\toprule
						GM-LVeG-S & \textbf{91.24} & \textbf{91.38} & \textbf{41.51} & \textbf{91.02} & 39.24 \\
						\bottomrule
		\end{tabular}}}}
		\caption{\label{tab:parse_acc}Parsing accuracy on the testing data of WSJ. EX indicates the exact match score.}
		\vskip -.0in
	\end{table}

\begin{figure}[!ht]
		\vskip -.0in
		\begin{minipage}[ht]{\linewidth} 
			\centering 
			\includegraphics[width=\linewidth]{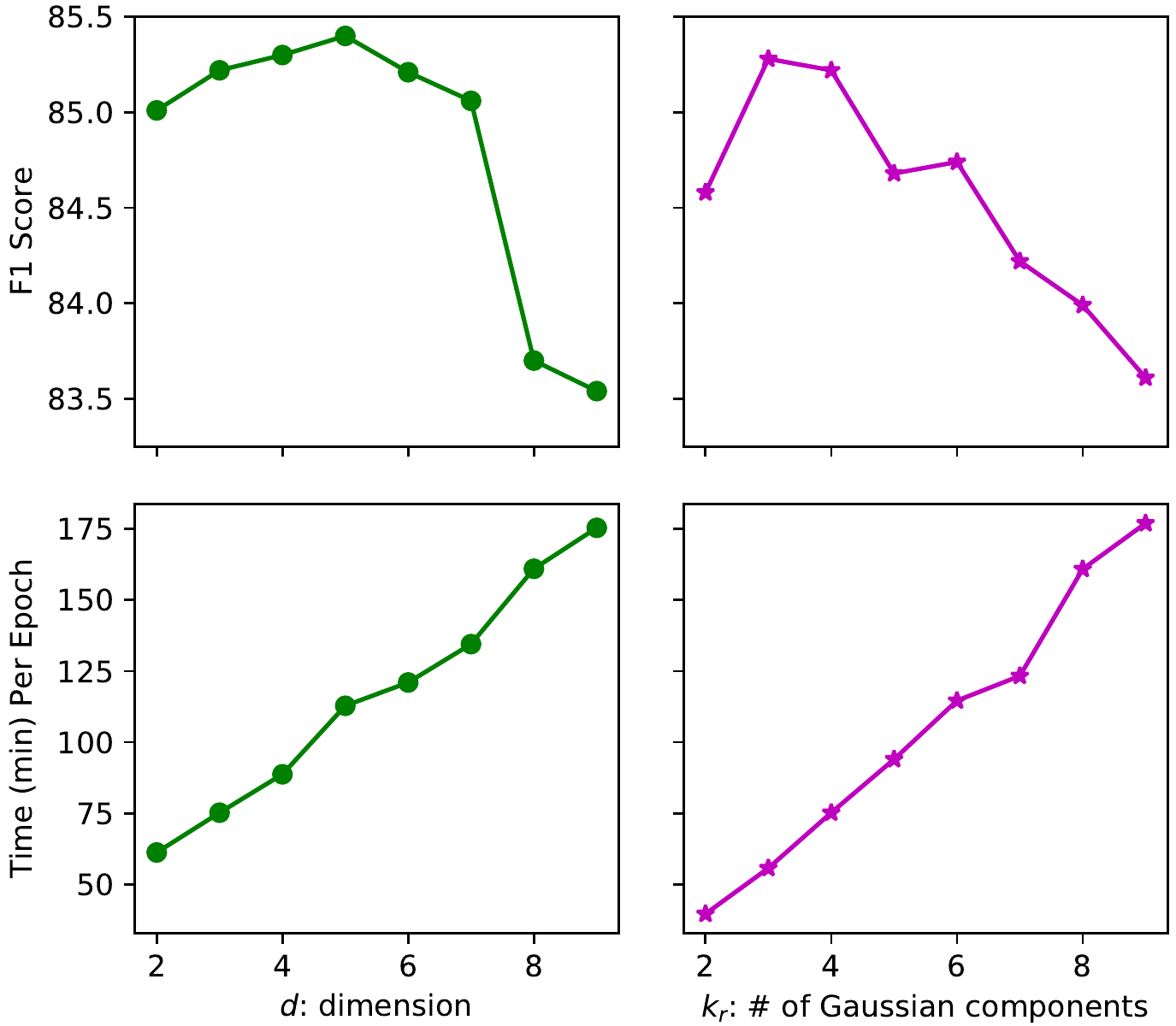}
			\vspace*{-1.5em} 
			\caption{\label{fig:dim_comp}F1 score and average time (min) consumed per epoch in learning. \textbf{Left}: \# of Gaussian components fixed to 4 with different dimensions; \textbf{Right}: dimension of Gaussians fixed to 3 with different \# of Gaussian components.} 
		\end{minipage} 
	\end{figure}

	We further investigate the influence of the latent vector dimension and the Gaussian component number on the efficiency and the parsing accuracy .
	We experiment on a small dataset (statistics of this dataset are in the supplemental material).
	We first fix the component number to 4 and experiment with the dimension 2, 3, 4, 5, 6, 7, 8, 9.
	Then we fix the dimension to 3 and experiment with the component number 2, 3, 4, 5, 6, 7, 8, 9.
	F1 scores on the development data are shown in the first row in Figure~\ref{fig:dim_comp}. 
	Average time consumed per epoch in learning is shown in the second row in Figure~\ref{fig:dim_comp}.
	When $K_r = 4$, the best dimension is 5;
	when $d = 3$, the best Gaussian component number is 3.
	A higher dimension or a larger Gaussian component number hurts the model performance and requires much more time for learning.
	Thus our choice of $K_r = 4$ and $d = 3$ in GM-LVeGs for parsing is a good balance between the efficiency and the parsing accuracy.
	
	%

	\section{Discussion}\label{sec:dis}

	It shall be noted that in this paper we choose to focus on the basic framework and algorithms of LVeGs, 
	and therefore we leave a few important extensions for future work.
	One extension is to incorporate contextual information of words and constituents. 
	which is a crucial technique that can be found in most state-of-the-art approaches to parsing or POS tagging.
	One possible way to utilize contextual information in LVeGs is to allow the words in the context of an anchored production rule to influence the rule's weight function.
	For example, we may learn neural networks to predict the parameters of the Gaussian mixture weight functions in a GM-LVeG from the pre-trained embeddings of the words in the context.
	

	In GM-LVeGs, we currently use the same number of Gaussian components for all the weight functions.
	A more desirable way would be automatically determining the number of Gaussian components for each production rule based on the ideal refinement granularity of the rule,
	e.g., we may need more Gaussian components for $\text{NP}\Ra \text{DT\,NN}$ than for $\text{NP}\Ra \text{DT\,JJ}$,
	since the latter is rarely used.
	There are a few possible ways to learn the component numbers such as greedy addition and removal, the split-merge method, and sparsity priors over mixture weights.

	An interesting extension beyond LVeGs is to have a single continuous space for subtypes of all the nonterminals.
	Ideally, subtypes of the same nonterminal or similar nonterminals are close to each other.
	The benefit is that similarity between nonterminals can now be modeled.
	%

	
	\section{Conclusion}
	We present Latent Vector Grammars (LVeGs) that associate each nonterminal with a latent continuous vector space representing the set of subtypes of the nonterminal.
	For each production rule, a LVeG defines a continuous weight function over the subtypes of the nonterminals involved in the rule.
	We show that LVeGs can subsume latent variable grammars and compositional vector grammars as special cases.
	We then propose Gaussian mixture LVeGs (GM-LVeGs).
	which formulate weight functions of production rules by mixtures of Gaussian distributions.
	The partition function and the expectations of fine-grained production rules in GM-LVeGs can be efficiently computed using dynamic programming,
	which makes learning and inference with GM-LVeGs feasible.
	We empirically show that GM-LVeGs can achieve competitive accuracies on POS tagging and constituency parsing.  
	
	%
	%
	%
	%
	%
	%

	\section*{Acknowledgments}
	This work was supported by the National Natural Science Foundation of China (61503248), Major Program of Science and Technology Commission Shanghai Municipal (17JC1404102), and Program of Shanghai Subject Chief Scientist (A type) (No.15XD1502900).
	We would like to thank the anonymous reviewers for their careful reading and useful comments.
	
	\bibliographystyle{acl_natbib}
	\bibliography{acl2018}
	
	\appendix
	
	
%
%
		
		\section*{\centering Supplementary Material}

		\section*{\centering Abstract}
		
		This supplementary material contains the following contents.
		(1) The derivation of the gradient formulation (Equation 12 in the paper).
		(2) The general idea of calculating analytic gradients for all the parameters in Gaussian Mixture Latent Vector Grammars (GM-LVeGs).
		(3) Algorithmic complexity and running time.
		(4) Statistics of the data used for part-of-speech (POS) tagging and constituency parsing.
		(5) Additional experimental results and analysis of GM-LVeGs.

		\section{Derivation of Gradient Formulations}	
		Given a training dataset $D = \{(T_i, \bw_i)\,|\, i = 1,\ldots,m\}$ containing $m$ samples,
		we minimize the negative log conditional likelihood during learning of GM-LVeGs:
		\begin{eqnarray}
		\label{eq:learn_obj_sup}
		\Cl(\Theta) = -\log \prod_{i = 1}^{m} P(T_{i} | \bw_i; \Theta)\,,
		\end{eqnarray}
		where $T_{i}$ is the gold parse tree with unrefined nonterminals for the sentence $\bw_i$,
		and $\Theta$ is the set of parameters in GM-LVeGs.

		We define $t$ as a parse tree with nonterminal subtypes,
		denote by $f_{r}(t)$ the number of occurrences of the unrefined rule $r$ in the unrefined parse tree 
		that is obtained by replacing all the subtypes in $t$ with the corresponding nonterminals,
		and use $\br$ to represent a fine-grained production rule of $r$, 
		which is represented by the concatenation of the latent vectors of the nonterminals in $r$. 
		
		The weight of $t$
		is defined as:
		\begin{align}
		s_{t} = \prod_{\br\in t} W_r(\br)\,.
		\end{align}
		The weight of $T$, a parse tree with unrefined nonterminals, is defined as:
		\begin{align}
		s_{T} = \int_{t\sim T} s_{t}\,dt\,,
		\end{align}
		where $t\sim T$ indicates that $t$ is a parse tree with nonterminal subtypes that can be converted into a parse tree $T$ by replacing its nonterminal subtypes with the corresponding nonterminals. The weight of a sentence $\bw$ is defined as:
		\begin{align}
		s_{\bw} = \int_{t\sim\bw} s_{t}\,dt\,,
		\end{align}
		where $t\sim \bw$ indicates that $t$ is a parse tree of $\bw$ with nonterminal subtypes. Thus the conditional probability density of $t$ given $T$ is
		\begin{align}
		P(t | T) = \frac{s_t}{s_T}\,,
		\end{align}
		the conditional probability density of $t$ given $\bw$ is
		\begin{align}
		P(t | \bw) = \frac{s_t}{s_{\bw}}\,,
		\end{align}
		and the conditional probability of $T$ given $\bw$ is
		\begin{align}
		P(T | \bw) = \frac{s_T}{s_{\bw}} = \int_{t\sim T} P(t | \bw)\,dt\,.
		\end{align}
		Therefore, we rewrite Equation~\ref{eq:learn_obj_sup} as
		\begin{align}
		\Cl(\Theta) = -\sum_{i = 1}^{m} \log \int_{t\sim T_i} P(t | \bw_i)\,dt\,.
		\end{align}
		The derivative of $\Cl(\Theta)$ with respect to $\Theta_r$, 
		where $r$ is an unrefined production rule, is calculated by Equation~\ref{eq:learn_derive} in Table~\ref{tab:grad_derive}.
		
		\begin{table*}[!ht]
			\centering
			{\setlength{\tabcolsep}{.0em}
				\begin{tabular}{c}
					\toprule 
					\begin{minipage}{\linewidth}
						\vspace{-.5em}
						\begin{eqnarray}
						\Cl'(\Theta) &=& 
						-\sum_{i = 1}^{m} \int_{t\sim T_i}
						\frac{\left( P(t | \bw_i)\right)'}{\int_{t'\sim T_i} P(t' |\bw_i) \,dt' } \,dt \nonumber\\ 
						&=& -\sum_{i = 1}^{m} \int_{t\sim T_i}
						\frac{\left( P(t | \bw_i)\right)'}{\int_{t'\sim T_i} P(t' |\bw_i) \,dt' } \times \frac{P(t | \bw_i)}{P(t | \bw_i)} \,dt \nonumber\\
						&=& -\sum_{i = 1}^{m}  \int_{t\sim T_i}
						\frac{P(t | \bw_i)}{\int_{t'\sim T_i} P(t' |\bw_i) \,dt' } \times \left( \log P(t | \bw_i) \right)' \,dt \nonumber\\
						&=& -\sum_{i = 1}^{m}  \int_{t\sim T_i}
						P(t | T_i) \times \left( \log \frac{\prod_{\br\in t} W_r(\br)}{\int_{t'\sim\bw_i} \prod_{\br\in t'} W_r(\br) \,dt' } \right)' \,dt \nonumber\\
						&=& -\sum_{i = 1}^{m} \int_{t\sim T_i}
						P(t | T_i) \times \left(\log\prod_{\br\in t} W_r(\br) - \log\int_{t'\sim\bw_i} \prod_{\br\in t'} W_r(\br) \,dt' \right)' \,dt \nonumber\\
						&=& -\sum_{i = 1}^{m}\left( 
						\int_{t\sim T_i} P(t | T_i) \left(\sum_{\br\in t}\log W_r(\br) \right)' \,dt
						- \left(\log\int_{t'\sim\bw_i} \prod_{\br\in t'} W_r(\br) \,dt' \right)'
						\right) \nonumber\\
						&=&  -\sum_{i = 1}^{m}\left( 
						\int_{t\sim T_i} P(t | T_i) \sum_{\br\in t} \frac{W_r^{'}(\br)}{W_r(\br)} \,dt
						- \int_{t'\sim\bw_i} P(t' | \bw_i) \sum_{\br\in t'} \frac{W_r^{'}(\br)}{W_r(\br)} \,dt'
						\right) \nonumber\\
						\label{eq:learn_derive}
						&=& -\sum_{i = 1}^{m} \int_{\br} 
						\frac{\mathbb{E}_{P(t|T_i)}[f_r(t)] -\mathbb{E}_{P(t|\bw_i)}[f_r(t)]}{W_r(\br)} \times W_r^{'}(\br) \,d\br\,.
						\end{eqnarray}
					\end{minipage}\\
					\toprule
					\begin{minipage}{\linewidth}
						\begin{eqnarray}
						\label{eq:grad_dmw_final}
						\frac{\partial\Cl(\Theta)}{\partial\rho_{r, k}} 
						&=&\sum_{i = 1}^{m} \int_{\br} \psi(\br) \cdot \Cn_{r, k}(\br)\,d\br\,. \\
						\label{eq:grad_dmu_final}
						\frac{\partial\Cl(\Theta)}{\partial\bmu_{r, k}} 
						&=&\sum_{i = 1}^{m} \int_{\br} \psi(\br) \cdot \Cn_{r, k}(\br) \cdot \rho_{r, k} 
						\bsi_{r, k}^{-1}  (\br - \bmu_{r, k})\,d\br\,. \\
						\label{eq:grad_dsig_final}
						\frac{\partial\Cl(\Theta)}{\partial\bsi_{r, k}} 
						&=&\sum_{i = 1}^{m} \int_{\br} \psi(\br) \cdot \Cn_{r, k}(\br) \cdot \rho_{r, k}
						\bsi_{r, k}^{-1} \frac{(\br - \bmu_{r, k}) (\br - \bmu_{r, k})^{T} \bsi_{r, k}^{-1} - I}{2}\,d\br\,.
						\end{eqnarray}
					\end{minipage}\\
					\bottomrule
			\end{tabular}}
			\caption{\label{tab:grad_derive} Derivation of gradient formulations.}
		\end{table*}
		
		\section{Calculation of Analytic Gradients}
		In GM-LVeGs, $\Theta_r$ is the set of parameters in a Gaussian mixture with $K_r$ mixture components:
		\begin{align}
		\Theta_r = \{(\rho_{r, k}, \bmu_{r, k}, \bsi_{r, k}) | k = 1,\ldots,K_r\}\,.
		\end{align}
		According to Equation~\ref{eq:learn_derive} in Table~\ref{tab:grad_derive}, we need to take derivatives of $W_r(\br)$ with respect to $\rho_{r, k}$, $\bmu_{r, k}$, and $\bsi_{r, k}$ respectively:
		\begin{align}
		\label{eq:grad_dmw_sup}
		\!\!\!\!\partial W_r(\br) / \partial\rho_{r, k} &= \Cn_{r, k}(\br) \,, \\
		\label{eq:grad_dmu_sup}
		\!\!\!\!\partial W_r(\br) / \partial\bmu_{r, k} &= \rho_{r, k}\Cn_{r, k}(\br)  \bsi_{r, k}^{-1}  (\br - \bmu_{r, k}) \,,\!\!\!\! \\
		\label{eq:grad_dsig_sup}
		\!\!\!\!\partial W_r(\br) / \partial\bsi_{r, k} &= \rho_{r, k}\Cn_{r, k}(\br)  \bsi_{r, k}^{-1} \frac{1}{2} \left( \vphantom{ \bsi_{r, k}^{-1} } -I \right. \\
		&+ \left. (\br - \bmu_{r, k}) (\br - \bmu_{r, k})^{T} \bsi_{r, k}^{-1} \right)\,. \nonumber
		\end{align}
		For brevity, we define
		\begin{align}\label{eq:psi}
		\psi(\br) 
		= \frac{\mathbb{E}_{P(t|\bw_i)}[f_r(t)] - \mathbb{E}_{P(t|T_i)}[f_r(t)]}{W_r(\br)}\,.
		\end{align}
		Substituting Equations~\ref{eq:grad_dmw_sup}--\ref{eq:grad_dsig_sup} into Equation~\ref{eq:learn_derive},
		we have the full gradient formulations of all the parameters (Equations~\ref{eq:grad_dmw_final}--\ref{eq:grad_dsig_final} in Table~\ref{tab:grad_derive}).
		
		In the following discussion, we assume that all the Gaussians are diagonal.
		It can be verified that $\psi(\br)$ in Equation~\ref{eq:psi} is in fact a mixture of Gaussians,
		so multiplying $\psi(\br)$ by $\Cn_{r, k}(\br)$ in Equation~\ref{eq:grad_dmw_final}--\ref{eq:grad_dsig_final} results in another mixture of Gaussians.
		Below we consider the special case where the resulting Gaussian mixture contains only a single component:
		\begin{align}
		\psi(\br)\cdot\Cn_{r, k}(\br) = \lambda\cdot\Cn(\br) \,.
		\end{align}
		Owing to the sum rule in integral, we can easily extend our derivation on the special case to the general case in which the Gaussian mixture contains multiple components.
		Because $\Cn(\br)$ is diagonal, it can be factorized as:
		\begin{align}
		\Cn(\br) = \Cn(\br^1)\times\cdots\times\Cn(\br^{|\br|})\,,
		\end{align}
		where $\Cn(\br^1),\ldots,\Cn(\br^{|\br|})$ are univariate Gaussians (or normal distributions), $\br^{d}$ ($1\le d\le |\br|$) refers to the $d$-th element of $\br$, and $|\br|$ is the dimension of $\br$.
		The integral in Equation~\ref{eq:grad_dmw_final} can be readily calculated as
		\begin{align}
		\lambda \cdot \int \Cn(\br) \,d\br = \lambda\,.
		\end{align}
		For Equation~\ref{eq:grad_dmu_final}, consider taking the derivative with respect to the mean in dimension $d$. Since the means in different dimensions are independent, to solve the integral in Equation~\ref{eq:grad_dmu_final}, we only need to solve the following integral:
		\begin{align}
		&\lambda\cdot\int \Cn(\br) \br^{d}\, d\br^{1}\ldots \,d\br^{|\br|} \nonumber \\
		&=\lambda\cdot\int \Cn(\br^1)\,d\br^{1} \times \cdots \times \int\Cn(\br^d)\br^d\,d\br^{d}  \nonumber \\
		&\times \ldots   \times\int \Cn(\br^{|\br|})\, d\br^{|\br|} \nonumber \\
		&=\lambda\cdot\int\Cn(\br^d)\br^d\,d\br^{d}\,. \label{eq:1st-order}
		\end{align}
		For Equation~\ref{eq:grad_dsig_final}, when taking the derivative with respect to the variance in dimension $d$, we also need to solve Equation~\ref{eq:1st-order} and additionally need to solve the following integral:
		\begin{align}
		&\lambda\cdot\int \Cn(\br) \br^{d} \br^{d}\, d\br^{1}\ldots \,d\br^{|\br|} \nonumber \\
		&=\lambda\cdot\int \Cn(\br^1)\,d\br^{1} \times \cdots \times \int\Cn(\br^d)\br^d\br^d\,d\br^{d}  \nonumber \\
		&\times \ldots   \times\int \Cn(\br^{|\br|})\, d\br^{|\br|} \nonumber \\
		&=\lambda\cdot\int\Cn(\br^d)\br^d\br^d\,d\br^{d}\,. \label{eq:2nd-order}
		\end{align}
		
		The integrals in Equation~\ref{eq:1st-order} and Equation~\ref{eq:2nd-order} are the first order moment and the second order moment of the univariate Gaussian $\Cn(\br^d)$ respectively, and both of them can be calculated exactly. Therefore, we can calculate analytic gradients of all the parameters in GM-LVeGs.

		\section{Algorithmic Complexity and Running Time}
		
		The time complexity of the learning algorithm for each sentence in each epoch is approximately $\mathcal{O}(cn^3kmd + cklmd^2)$. The first term is the time complexity of the extended inside-outside algorithm, where $c$ is the number of binary productions in CNF, $n$ is the length of the sentence, $k$ is the Gaussian component number of each rule weight function, $m$ is the maximum Gaussian component number of an inside or outside score after pruning, and $d$ is the dimension of diagonal Gaussians. $kmd$ is generally much smaller than $cn^3$. It shall be noted that $m$ is bounded by $k_{max}$, which is set to 50 in our experiments. The second term is the approximate time complexity of gradient calculation, where $l$ is the number of times that a production rule is used in all possible parses of a sentence. The second term is much smaller than the first term in general.
		
		We run our learning and inference algorithms with a CPU cluster without any GPU.
		In our POS tagging experiments,
		GM-LVeGs are only slightly slower than the baseline LVG models, and we can perform all the tagging experiments on all the datasets with our model within one day.
		For parsing, there is a trade-off between running time and parsing accuracy based on the amount of pruning. 
		For the best parsing accuracy of GM-LVeGs, it takes two weeks for training. 
		However, once we complete training, parsing can be done within three minutes on the whole testing data of WSJ.
		
		There are a few ways to improve the training efficiency. We currently use CPU parallelization at the sentence level in training, 
		but in the future we may take the advantage of GPU parallelization, e.g., we can vectorize the inside-outside algorithm for a batch of sentences of the same length. 
		Besides, for each long sentence, we can parallelize the inside or outside computation at the same recursive depth.
		
		\section{Data Statistics}\label{sec:data}
		
		\begin{table}[!ht]
			\centering
			\vskip -.0in
			{\setlength{\tabcolsep}{1.em}
				\makebox[\linewidth]{\resizebox{\linewidth}{!}{%
						\begin{tabular}{ccccc}
							\toprule
							\multirow{3}{*}{Dataset} & \multirow{3}{*}{\# of tokens} & \multicolumn{3}{c}{\# of sentences} \\
							\cmidrule{3-5}
							& & train & test & dev \\
							\toprule
							WSJ & 950028 & 39832 & 2416 & 1700\\
							\bottomrule
			\end{tabular}}}}
			\caption{\label{tab:data_parsing}Statistics of WSJ used for constituency parsing.}
			\vskip -.0in
		\end{table}	
		
		\begin{table*}[!ht]
			\centering
			\vskip -.0in
			{\setlength{\tabcolsep}{.8em}
				\makebox[\linewidth]{\resizebox{\linewidth}{!}{%
						\begin{tabular}{crrrrrrrrr}
							\toprule
							& WSJ & English & French & German & Russian & Spanish & Indonesian & Finnish & Italian \\
							\toprule
							\# of tokens & 1173766 & 254830 & 402197 & 298242 & 99389& 431587 & 121923 & 181022 & 292471\\
							train        &   38219 &  12543 &  14554 &  14118 & 4029 &  14187 &   4477 &  12217 &  12837\\
							dev          &    5462 &   2002 &   1596 &    799 &  502 &   1552 &    559 &    716 &    489\\
							test         &    5527 &   2077 &    298 &    977 &  499 &    274 &    557 &    648 &    489\\
							\bottomrule
			\end{tabular}}}}
			\caption{\label{tab:tag_data}Statistics of WSJ and UD (English, French, German, Russian, Spanish,  Indonesian, Finnish, and Italian treebanks). Numbers in the rows of train, test, and dev indicate the number of sentences in training, testing, and development data respectively.}
			\vskip -.0in
		\end{table*}	
		
		\begin{table*}[!ht]
			\centering
			\vskip -.0in
			{\setlength{\tabcolsep}{.3em}
				\makebox[\linewidth]{\resizebox{\linewidth}{!}{%
						\begin{tabular}{lcccccccccccccccccc}
							\toprule
							\multirow{3}{*}{Model} &
							\multicolumn{2}{c}{WSJ}  &
							\multicolumn{2}{c}{English}  &
							\multicolumn{2}{c}{French} &
							\multicolumn{2}{c}{German} &
							\multicolumn{2}{c}{Russian} &
							\multicolumn{2}{c}{Spanish} &
							\multicolumn{2}{c}{Indonesian} &
							\multicolumn{2}{c}{Finnish} &
							\multicolumn{2}{c}{Italian} \\ 
							\cmidrule(lr){2-3} \cmidrule(lr){4-5} \cmidrule(lr){6-7} \cmidrule(lr){8-9} \cmidrule(lr){10-11} \cmidrule(lr){12-13} \cmidrule(lr){14-15} \cmidrule(lr){16-17} \cmidrule(lr){18-19} 
							&
							\multicolumn{1}{c}{T}  & \multicolumn{1}{c}{S} &
							\multicolumn{1}{c}{T}  & \multicolumn{1}{c}{S} &
							\multicolumn{1}{c}{T}  & \multicolumn{1}{c}{S} &
							\multicolumn{1}{c}{T}  & \multicolumn{1}{c}{S} &
							\multicolumn{1}{c}{T}  & \multicolumn{1}{c}{S} &
							\multicolumn{1}{c}{T}  & \multicolumn{1}{c}{S} &
							\multicolumn{1}{c}{T}  & \multicolumn{1}{c}{S} &
							\multicolumn{1}{c}{T}  & \multicolumn{1}{c}{S} &
							\multicolumn{1}{c}{T}  & \multicolumn{1}{c}{S} \\
							\midrule
							LVG-D-1 & 96.50& 48.04& 91.80& 50.79& 93.55& 30.20& 86.52& 16.99& 81.21&  9.24& 91.79& 22.63& 89.08& 18.85& 83.15& 16.82& 94.00& 37.42\\
							LVG-D-2 & 96.57& 47.60& 92.17& 52.05& 93.86& 33.56& 86.93& 18.32& 81.46& 10.04& 92.10& 24.82& 89.16& 19.21& 83.34& 18.52& 94.45& 40.90\\
							LVG-D-4 & 96.57& 48.76& 92.30& 52.34& 93.96& 34.90& 87.18& 19.86& 81.95& 11.85& 92.37& 24.82& 89.28& 19.57& 83.76& 18.83& 94.60& 42.54\\
							LVG-D-8 & 96.60& 49.14& 92.31& 53.06& 93.78& 34.90& 87.52& 21.60& 81.54& 11.25& 92.26& 23.72& 89.23& 19.39& 83.68& 18.67& 94.70& 42.95\\
							LVG-D-16& 96.62& 48.74& 92.31& 52.67& 93.75& 34.90& 87.38& 20.98& 81.91& 12.25& 92.47& 24.82& 89.27& 20.29& 83.81& 19.29& 94.81& 45.19\\
							\toprule
							LVG-G-1 & 96.11& 43.68& 90.84& 44.92& 92.69& 26.51& 86.71& 17.40& 81.22& 10.22& 91.85& 22.63& 88.93& 18.31& 82.94& 16.36& 93.64& 33.74\\
							LVG-G-2 & 96.27& 45.57& 92.11& 51.37& 93.28& 28.19& 87.87& 19.86& 81.51& 11.45& 92.29& 23.36& 89.19& 18.49& 83.29& 17.44& 94.20& 38.45\\
							LVG-G-4 & 96.50& 48.19& 92.90& 54.31& 94.06& 32.55& 88.31& 20.78& 82.64& 11.85& 92.58& 24.45& 89.58& 19.03& 83.76& 19.44& 95.00& 45.40\\
							LVG-G-8 & 96.76& 50.38& 93.29& 56.67& 94.57& 37.25& 88.75& 21.70& 82.85& 14.86& 92.95& 29.20& 89.78& 20.29& 84.69& 21.76& 95.42& 46.83\\
							LVG-G-16& 96.78& 50.88& 93.30& 57.54& 94.52& 34.90& 88.92& 24.05& 84.03& 16.63& 93.21& 27.37& 90.09& 21.19& 85.01& 20.53& 95.46& 48.26\\
							\toprule
							GM-LVeG-D& 96.99& 53.10& \textbf{93.66}& \textbf{59.46}& 94.73& \textbf{39.60}& 89.11& 24.77& \textbf{84.21}& 17.84& \textbf{93.76}& \textbf{32.48}& \textbf{90.24}& \textbf{21.72}& 85.27& \textbf{23.30}& 95.61& \textbf{50.72}\\
							GM-LVeG-S& \textbf{97.00}& \textbf{53.11}& 93.55& 58.11& \textbf{94.74}& 39.26& \textbf{89.14}& \textbf{25.58}& 84.06& \textbf{18.44}& 93.52& 30.66& 90.12& \textbf{21.72}& \textbf{85.35}& 22.07& \textbf{95.62}& 49.69\\
							\bottomrule
			\end{tabular}}}}
			\caption{\label{tab:tag_acc_sup}Token accuracy (T) and sentence accuracy (S) for POS tagging on the testing data. The numerical postfix of each LVG model indicates the number of nonterminal subtypes, and hence LVG-G-1 denotes HMM.}
			\vskip -.0in
		\end{table*}
		
		Statistics of the data used for constituency parsing are shown in Table~\ref{tab:data_parsing}.
		Statistics of the data used for POS tagging are summarized in Table~\ref{tab:tag_data}.
		In the experiments of constituency parsing, 
		in order to study the influence of the dimension of Gaussians and the number of Gaussian components on the parsing accuracy,
		we experimented on a small dataset.
		The small dataset only contains section 4 and section 5 of WSJ.
		In the two sections, we use file IDs from 80-89 in the two sections are used for testing, 90-99 for development, and the rest are used for training.
		The resulting dataset contains 3599 training samples, 426 test samples, and 375 development samples.
		
		\section{Additional Experimental Results}
		
		The complete experimental results of POS tagging are shown in Table~\ref{tab:tag_acc_sup}. 
		In addition to the results shown in the paper, this table includes the tagging results of LVGs with 1, 2, 4, 8 subtypes for each nonterminal. 
		
\begin{figure}[h!]
		\vspace*{-.em} 
		\begin{minipage}{\linewidth} 
			\centering 
			\includegraphics[width=\linewidth]{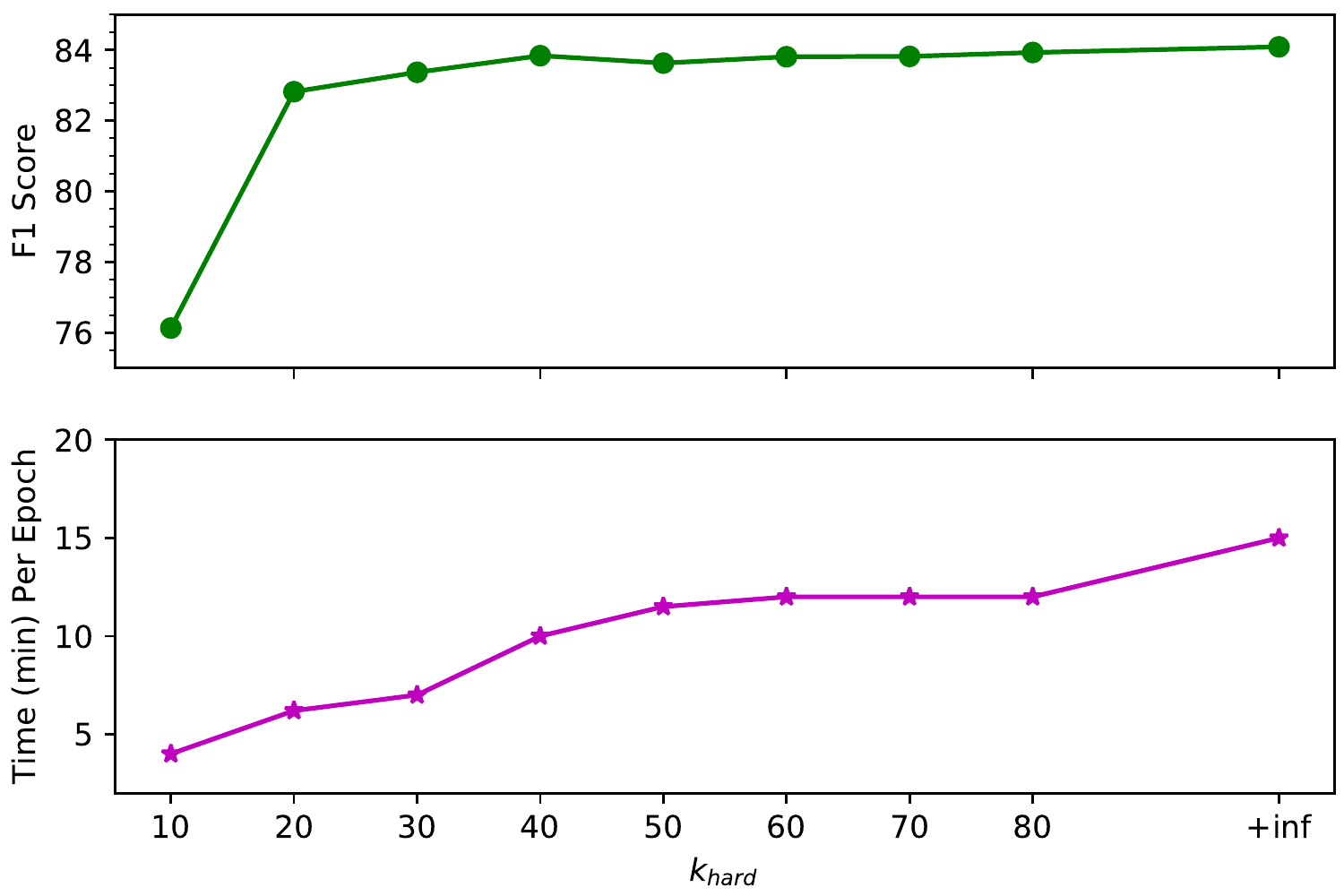}
			\caption{\label{fig:comp_prune}\textbf{Upper}: F1 scores with different $k_{hard}$; \textbf{Lower}: time (min) per epoch in learning with different $k_{hard}$.} 
		\end{minipage} 
		\vspace*{-1.5em}
		\end{figure}

		In the experiments of constituency parsing, 
		in order to investigate the impact of the maximum Gaussian component number of inside and outside scores,
		we experiment with a new pruning technique.
		Specifically, we use a maximum component-pruning threshold $k_{hard}$. We do not prune any Gaussian component if an inside or outside score has no more than $k_{hard}$ Gaussian components; otherwise we keep only $k_{hard}$ Gaussian components with the largest mixture weights.
		We experiment on the small dataset mentioned in Section~\ref{sec:data}.
		For efficiency, we train GM-LVeG-D only on sentences of no more than 20 words 
		and test GM-LVeG-D only on testing sentences of no more than 25 words.
		We experiment with $k_{hard} = 10, 20, 30, 40, 50, 60, 70, 80$. 
		The results are shown in Figure~\ref{fig:comp_prune}.
		We also experiment without component pruning, which corresponds to the rightmost point in Figure~\ref{fig:comp_prune}.

		We can see that a weaker component pruning or a larger $k_{hard}$ results in a better F1 score.
		However, it takes much more time per epoch for learning, as is shown in the lower figure in Figure~\ref{fig:comp_prune}.
		We find that $k_{hard} = 40$ produces a good F1 score and also admits efficient learning. 
		Therefore, in the experiments of constituency parsing, we use $k_{min} = 40$ in learning for the component-pruning technique introduced in Section 3.2 in the paper.

\end{document}